\documentclass{article}
\usepackage{spconf,amsmath,graphicx}
\usepackage{booktabs}
\usepackage{stfloats}
\usepackage{multirow}
\usepackage{amsfonts}
\usepackage{enumitem}
\usepackage{setspace}

\setlist{nosep, leftmargin=14pt}

\usepackage{mwe} 

\title{Multimodal Self-supervised Learning for Lesion Localization}
\name{%
\begin{tabular}{c}
    {Hao Yang$^{1,2,3}$\qquad Hong-Yu Zhou$^{4}$\qquad Cheng Li$^{1}$\qquad Weijian Huang$^{1,2,3}$}\qquad Jiarun Liu$^{1,2,3}$ \\
    {Yong Liang$^{2,5}$\qquad Guangming Shi$^{2}$\qquad Hairong Zheng$^{1}$\qquad Qiegen Liu$^{6}$\qquad Shanshan Wang$^{1,\dagger}$\thanks{$^{\dagger}$ Corresponding author. ss.wang@siat.ac.cn}}
  \end{tabular}
}
\address{$^{1}$Paul C. Lauterbur Research Center for Biomedical Imaging, \\
Shenzhen Institute of Advanced Technology, Chinese Academy of Sciences, Shenzhen, China\\
$^{2}$Peng Cheng Laboratory, Shenzhen, China\\
$^{3}$University of Chinese Academy of Sciences, Beijing, China\\
$^{4}$Department of Computer Science, The University of Hong Kong, Pokfulam, China\\
$^{5}$Pazhou Laboratory (Huangpu), Guangzhou, China \\
$^{6}$Department of Electronic Information Engineering, Nanchang University, China
}

\begin{document}
%
\maketitle
\begin{abstract}
Multimodal deep learning utilizing imaging and diagnostic reports has made impressive progress in the field of medical imaging diagnostics, demonstrating a particularly strong capability for auxiliary diagnosis in cases where sufficient annotation information is lacking. Nonetheless, localizing diseases accurately without detailed positional annotations remains a challenge. Although existing methods have attempted to utilize local information to achieve fine-grained semantic alignment, their capability in extracting the fine-grained semantics of the comprehensive context within reports is limited. To address this problem, a new method is introduced that takes full sentences from textual reports as the basic units for local semantic alignment. This approach combines chest X-ray images with their corresponding textual reports, performing contrastive learning at both global and local levels. The leading results obtained by this method on multiple datasets confirm its efficacy in the task of lesion localization.
\end{abstract}
\section{Introduction}
\label{sec:intro}

Deep learning technologies have made remarkable progress in the medical field, especially in medical imaging diagnostics \cite{titano2018automated,wang2021annotation,huang2023enhancing,zhou2023transformer,zhou2023unified}. However, these technologies also face challenges, as they are highly dependent on a large amount of precisely annotated data, whereas the process of data annotation is both time-consuming and labor-intensive \cite{wang2021annotation}. For some emerging or rare diseases, obtaining sufficient annotated data is extremely difficult. Besides, for disease categories not present during training, the model may require extensive adjustments or fine-tuning. To overcome these challenges, the use of image reports enriched with medical knowledge as learning resources has become particularly crucial \cite{zhou2023advancing,wu2023medklip,huang2021gloria}. These reports provide detailed descriptions of diseases, allowing deep learning models to autonomously learn medical features without the need for explicit annotations. Existing relevant studies often employ multimodal deep learning techniques, combining medical images with corresponding text reports. The trained model can be applied to classify samples that were not explicitly annotated during the training phase. Studies have shown that this multimodal learning strategy can yield accurate results in certain disease classification tasks, with performance levels comparable to those of medical experts.

\begin{figure}[bht]
  \centering
  \centerline{\includegraphics[width=7.5cm]{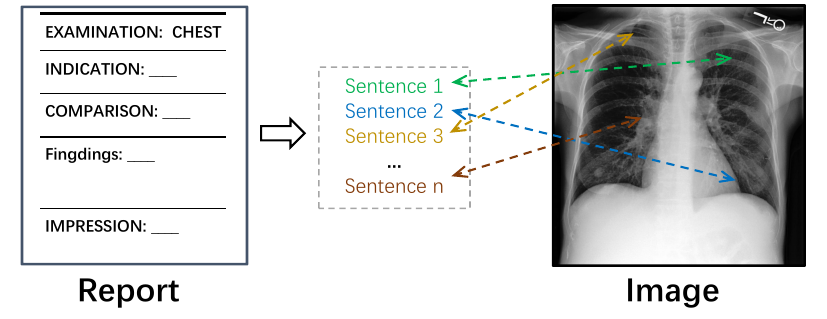}}
  \caption{The description in the diagnostic report typically corresponds to the findings in the imaging.}
  \medskip
\end{figure}

In medical research and practice, accurate classification of diseases is just the initial step in diagnosis; more crucial is the precise localization of the disease. High-precision localization capabilities of models not only enhance doctors' confidence in them but can also reduce biases and errors in the diagnostic process \cite{saporta2022benchmarking}. However, in deep learning research based on medical imaging reports, precise localization of diseases remains a significant challenge. This challenge primarily arises from the clear absence of location-level labels. Current research tends to rely on global features for implementing contrastive learning \cite{zhang2022contrastive,wu2023medklip}, but this method often lacks semantic granularity, thus overlooking detailed information. To overcome this problem, some research methods have been improved by aligning words with local regions of the image \cite{huang2021gloria,boecking2022making}. However, since the meanings of words vary in different contexts and circumstances, the semantic scope they represent is often limited and cannot precisely capture the complete meaning of pathological descriptions. Furthermore, local features in images typically reveal characteristics directly related to pathology, such as shape, size, and texture. Using words alone to align with these features is inappropriate. As shown in Fig. 1, complete sentences can often match the local features of an image more accurately and demonstrate richer semantic connections. Currently, deep learning methods based on medical imaging reports still face issues of insufficient precision and accuracy in disease localization, prompting the need for more in-depth exploration and research.

In this paper, we propose a self-supervised learning approach that eliminates the dependence on costly bounding box annotations and achieves precise localization of lesions in chest X-ray images using only concise lesion descriptions. As illustrated in Figure 2, we align the global features of medical reports with the global features of images. Simultaneously, we align sentences, as complete semantic units, with the local features of images, learning shared latent semantics. It is important to note that while the contrastive loss between local and global features has been explored in previous works \cite{huang2021gloria}, our main innovation lies in training our model using sentence-based embeddings rather than traditional word-based embeddings. Experimental results demonstrate that our method exhibits robust results in disease localization, even when faced with unseen diseases. Moreover, our approach still significantly outperforms the current state-of-the-art methods, further validating the superiority of our method.

\section{Method}
\label{sec:format}

\subsection{Image Encoder}

In this article, to ensure fairness in evaluation, we adopted the same architecture that is prevalent in current research, namely the ResNet-50 \cite{he2016deep}, as the backbone of our visual encoder \(E_v\). When processing an input image \(x_v\), we extract image features from specific intermediate convolutional layers using the visual encoder. Additionally, we extract features from the last convolutional layer of ResNet-50 and employ average pooling to obtain the global features. Subsequently, both the local features \(v_l\) and the global features \(v_g\) undergo dimensionality adjustment using \(1\times1\) convolutions and linear layers to match the dimensionality of the text features. Ultimately, we obtain the local features \(v_l\in \mathbb{R}^{D\times M}\) and the global features \(v_g\in \mathbb{R}^D\), where \(M\) denotes the number of sub-regions in the local features, and \(D\) represents the feature dimension.

\begin{figure}[bht]
  \centering
  \centerline{\includegraphics[width=8cm]{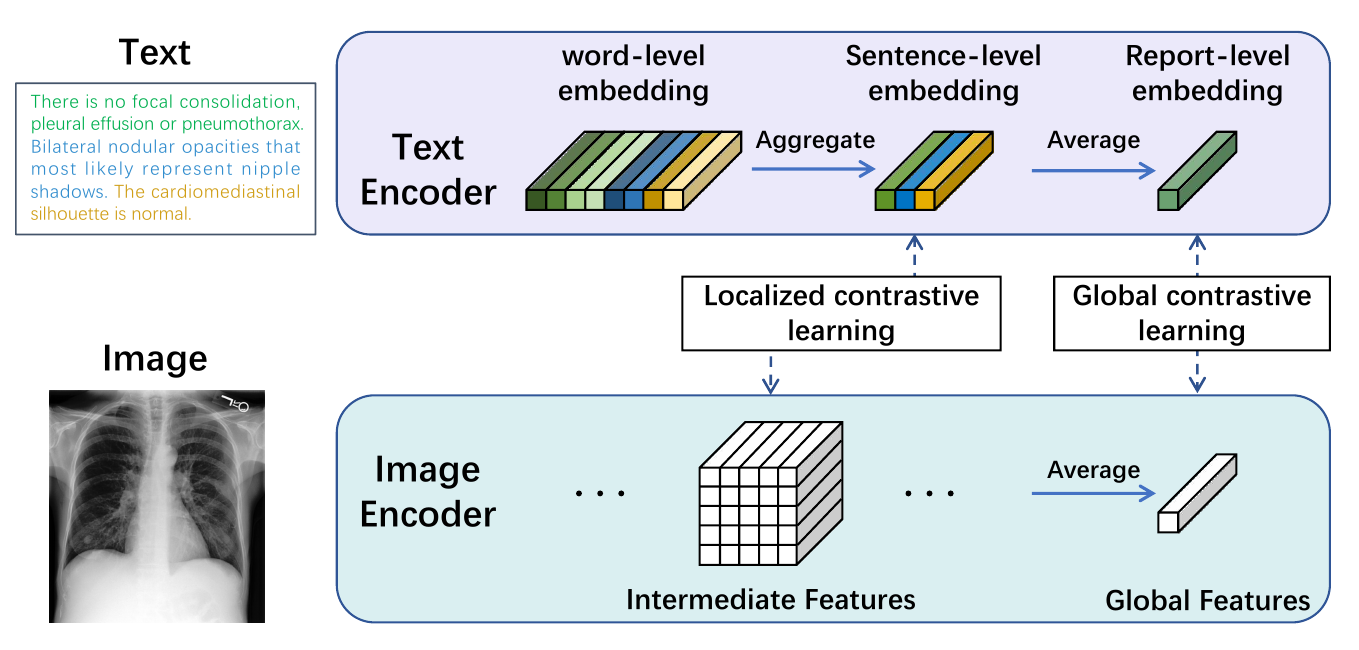}}

  \caption{The framework is based on global and local contrastive learning. Global contrastive learning uses global image features and text features. Local contrastive learning aligns sentence-level features with local image features.}
  \medskip
\end{figure}

\subsection{Text Encoder}

The original radiology reports usually comprise a 'Findings' section, which details clinical observations, and an 'Impressions' section, summarizing the clinical assessment. These sections typically contain multiple sentences. While existing methods \cite{huang2021gloria,boecking2022making} often treat words as separate entities, it's important to note that words can carry different meanings in different contexts. Treating words as independent entities can sometimes lead to erroneous semantic interpretations. Therefore, the approach proposed in this paper considers all words in a sentence collectively to capture and express more comprehensive and precise semantic content.

To obtain text embeddings, we utilize BioClinicalBERT \cite{alsentzer2019publicly} as the encoder. When processing a medical report \( x_t \) that contains \( Q \) words and \( P \) sentences, we first perform subword tokenization on each word to obtain \( q_i \) subwords, while also identifying the number of \( p_i \) words contained in each sentence. Our tokenizer generates a total of \(N = {\textstyle \sum_{i=1}^{Q} q_i} \) word piece embeddings as input for the text encoder.

The text encoder then extracts features for each word piece independently, resulting in an embedding output represented as \( t_t\in \mathbb{R}^{D\times N} \), where \( D \) is the dimension of the word piece feature embedding. Word-level embeddings, represented as \( t_{wj} = {\textstyle \sum_{i=1}^{q_j} t_{ti}} \), are obtained by aggregating the corresponding word piece embeddings; similarly, sentence-level embeddings, represented as \( t_{sj} = {\textstyle \sum_{i=1}^{p_j} t_{wi}} \), are obtained by aggregating all word-level embeddings within each sentence. On this basis, we define a report-level embedding \( t_g = \frac{1}{P} {\textstyle \sum_{i=1}^{P} t_{si}} \), which is a high-dimensional representation of a comprehensive semantic understanding of the entire report content.

\subsection{Global and Local contrastive learning}

Given that medical reports typically provide detailed descriptions corresponding to observations in medical images, we anticipate that images and their corresponding reports will exhibit consistent semantic attributes within a multimodal feature space. Drawing inspiration from methods in the literature \cite{huang2021gloria}, we applied a strategy for aligning global and local semantics, As shown in Fig. 2. In the context of global contrastive learning, the use of positive and negative pairs is aimed at training models to distinguish between matched (positive) and unmatched (negative) sample pairs. Positive pairs consist of naturally associated samples, such as a medical image and its corresponding report, while negative pairs are randomly paired, unrelated samples. This approach enables models to learn and identify the key features that link images to their corresponding reports, facilitating the recognition and differentiation of these associations. The formal representation is as follows:
\begin{equation}
L_g^{(v|t)} = -\frac{1}{B} \sum_{i}^{B}\left ( log\frac{exp(v_g^i\cdot t_g^i)/\tau _1}{ {\textstyle \sum_{k=1}^{B}} exp(v_g^i\cdot t_g^k)/\tau _1} \right ) 
\end{equation}

\begin{equation}
L_g^{(t|v)} =-\frac{1}{B} \sum_{i}^{B}\left ( log\frac{exp(v_g^i\cdot t_g^i)/\tau _1}{ {\textstyle \sum_{k=1}^{B}} exp(v_g^k\cdot t_g^i)/\tau _1} \right ) 
\end{equation}
where \(\tau _1\) is a temperature parameter and \(B\) is the batch size.

During local semantic alignment, we identified challenges with word-level alignment, as the semantic granularity of words might be too specific, or some terms might lack substantive semantic content, preventing effective matching with the semantics associated with local regions in the image. To address this, we propose to use sentence-level representations rather than word-level representations for alignment. Based on this approach, we adopted a method of aligning sentence-level embeddings with the local features of the image. Following the local contrastive loss approach described in \cite{huang2021gloria}, the similarity matrix between S sentences and M image sub-regions is represented as:
\begin{equation}
s=v_l^Tt_s
\end{equation}

The image features representation \(c_i\) weighted by sentences is expressed as:
\begin{equation}
c_i= \sum_{j=1}^{M} log\frac{exp(s_{ij})/\tau _2}{ {\textstyle \sum_{k=1}^{M}} exp(s_{ik})/\tau _2}v_j
\end{equation}
where \(\tau _2\) is a temperature parameter. Using the local feature matching function \( Z \), aggregate all sentence features with the weighted sentence features:
\begin{equation}
Z(x_t,x_v)= log(\sum_{j=1}^{M} {exp(c_i\cdot t_{si})/\tau _3})^{\tau _3}
\end{equation}
where \(\tau _3\) is another scaling factor. The local contrastive loss is defined as:
\begin{equation}
L_l^{(v|t)} = -\frac{1}{B} \sum_{i}^{B}\left ( log\frac{exp(Z(x_{vi},x_{ti}))/\tau _1}{ {\textstyle \sum_{k=1}^{B}} exp(Z(x_{vi},x_{tk})/\tau _1} \right ) 
\end{equation}
\begin{equation}
L_l^{(t|v)} = -\frac{1}{B} \sum_{i}^{B}\left ( log\frac{exp(Z(x_{vi},x_{ti}))/\tau _1}{ {\textstyle \sum_{k=1}^{B}} exp(Z(x_{vk},x_{ti})/\tau _1} \right ) 
\end{equation}

Our learning framework jointly optimizes the global and local losses, with the final total loss being represented as:
\begin{equation}
L=L_g^{(v|t)}+L_g^{(t|v)}+L_l^{(v|t)}+L_l^{(t|v)}
\end{equation}

\begin{table}[ht]   
    \centering
    \caption{The Dice and IoU metrics were evaluated on the RSNA Pneumonia dataset and the Covid-Rural dataset, and they were compared to other state-of-the-art zero-shot learning region localization methods.}
    \fontsize{10}{11}\selectfont
    \label{t1}
    \begin{tabular}{ccccc}
        \toprule[1pt] 
        \multirow{3}{*}{Method}&\multicolumn{2}{c}{RSNA} &\multicolumn{2}{c}{Covid-19} \\  
        \cmidrule(lr){2-5}
                & IoU   & Dice  & IoU      & Dice  \\\midrule[1pt]
        {Gloria \cite{huang2021gloria}} & 0.218 & 0.347 & 0.065    & 0.114 \\
        {Biovil \cite{boecking2022making}} & 0.303 & 0.439 & 0.120    & 0.197 \\
        {MedKLIP \cite{wu2023medklip}} & 0.317 & 0.465 & 0.137    & 0.228 \\
        \textbf{Ours}   & \textbf{0.331} & \textbf{0.474} & \textbf{0.222}    & \textbf{0.336} \\
        \bottomrule[1pt]
    \end{tabular}
\end{table}

\begin{table*}[bht]
\centering
\caption{The Dice scores were evaluated on the MS-CXR dataset, and it was compared to other state-of-the-art zero-shot learning region localization methods.}
\fontsize{7.5}{10}\selectfont
\begin{tabular}{cccccccccc}
\toprule[1pt] 
Method & Pneumonia & Pneumothorax & Consolidation & Atelectasis & Edema & Cardiomegaly & Lung Opacity & Pleural Effusion & mean \\
{Gloria \cite{huang2021gloria}} & 0.417 & 0.181 & 0.443 & 0.442 & 0.315 & \textbf{0.567} & 0.298 & 0.476 & 0.392 \\
{Biovil \cite{boecking2022making}} & 0.472 & \textbf{0.217} & 0.433 & 0.405 & 0.326 & 0.560 & 0.294 & 0.352 & 0.382 \\
{MedKLIP \cite{wu2023medklip}} & 0.443 & 0.151 & 0.401 & 0.476 & \textbf{0.476} & 0.559 & 0.307 & 0.344 & 0.395 \\
\textbf{Ours}    & \textbf{0.576} & 0.163 & \textbf{0.538} & \textbf{0.538} & 0.433 & 0.485 & \textbf{0.468} & \textbf{0.525} & \textbf{0.466} \\
\bottomrule[1pt]
\end{tabular}
\end{table*}

\begin{table*}[bht]
\centering
\caption{The Intersection over Union (IoU) metric was evaluated on the MS-CXR dataset, and it was compared to other state-of-the-art zero-shot learning region localization methods.}
\fontsize{7.5}{10}\selectfont
\begin{tabular}{cccccccccc}
\toprule[1pt] 
Method & Pneumonia & Pneumothorax & Consolidation & Atelectasis & Edema & Cardiomegaly & Lung Opacity & Pleural Effusion & mean \\
{Gloria \cite{huang2021gloria}}  & 0.290 & 0.116 & 0.304 & 0.303 & 0.201 & \textbf{0.408} & 0.197 & 0.330 & 0.269 \\
{Biovil \cite{boecking2022making}}  & 0.328 & \textbf{0.137} & 0.297 & 0.275 & 0.213 & 0.406 & 0.188 & 0.224 & 0.259 \\
{MedKLIP \cite{wu2023medklip}} & 0.297 & 0.091 & 0.265 & 0.323 & \textbf{0.327} & 0.395 & 0.197 & 0.216 & 0.264 \\
\textbf{Ours}    & \textbf{0.425} & 0.106 & \textbf{0.386} & \textbf{0.388} & 0.294 & 0.330 & \textbf{0.325} & \textbf{0.368} & \textbf{0.328} \\
\bottomrule[1pt]
\end{tabular}
\end{table*}

\section{EXPERIMENTS AND RESULTS}
\label{sec:illust}

\subsection{Dataset}
The self-supervised methods were trained on the MIMIC-CXR \cite{johnson2019mimic} dataset, which is a publicly available dataset of chest X-ray images. The MIMIC-CXR dataset contains 377,110 images with each chest X-ray paired with its respective radiology report. We also use the RSNA Pneumonia \cite{shih2019augmenting}, COVID Rural \cite{desai2020chest}, and MS-CXR \cite{boecking2022making} datasets to evaluate the localization performance.

\subsection{Experimental details}
We performed hyperparameter optimization on the MIMIC-CXR validation dataset, which included batch size and learning rate. During this process, we utilized the Adam optimizer with an initial learning rate of 0.00002 and a momentum value of 0.9. Additionally, we implemented a learning rate decay mechanism that reduced the learning rate to 0.9 times its previous value after each epoch. The best model configuration from this hyperparameter sweep had a batch size of 128 and was trained for 4 epochs.
\subsection{Results}
We evaluated using the intersection over union (IoU) and Dice scores. Initially, we computed the cosine similarity between the projected phrase embeddings and the local image representations, resulting in a score grid within the range [-1, 1]. This similarity was later thresholded to calculate the IoU and Dice scores. The final results were defined as the average over the threshold values of [0.1, 0.2, 0.3, 0.4, 0.5].

\begin{figure}[bht]
  \centering
  \centerline{\includegraphics[width=8cm]{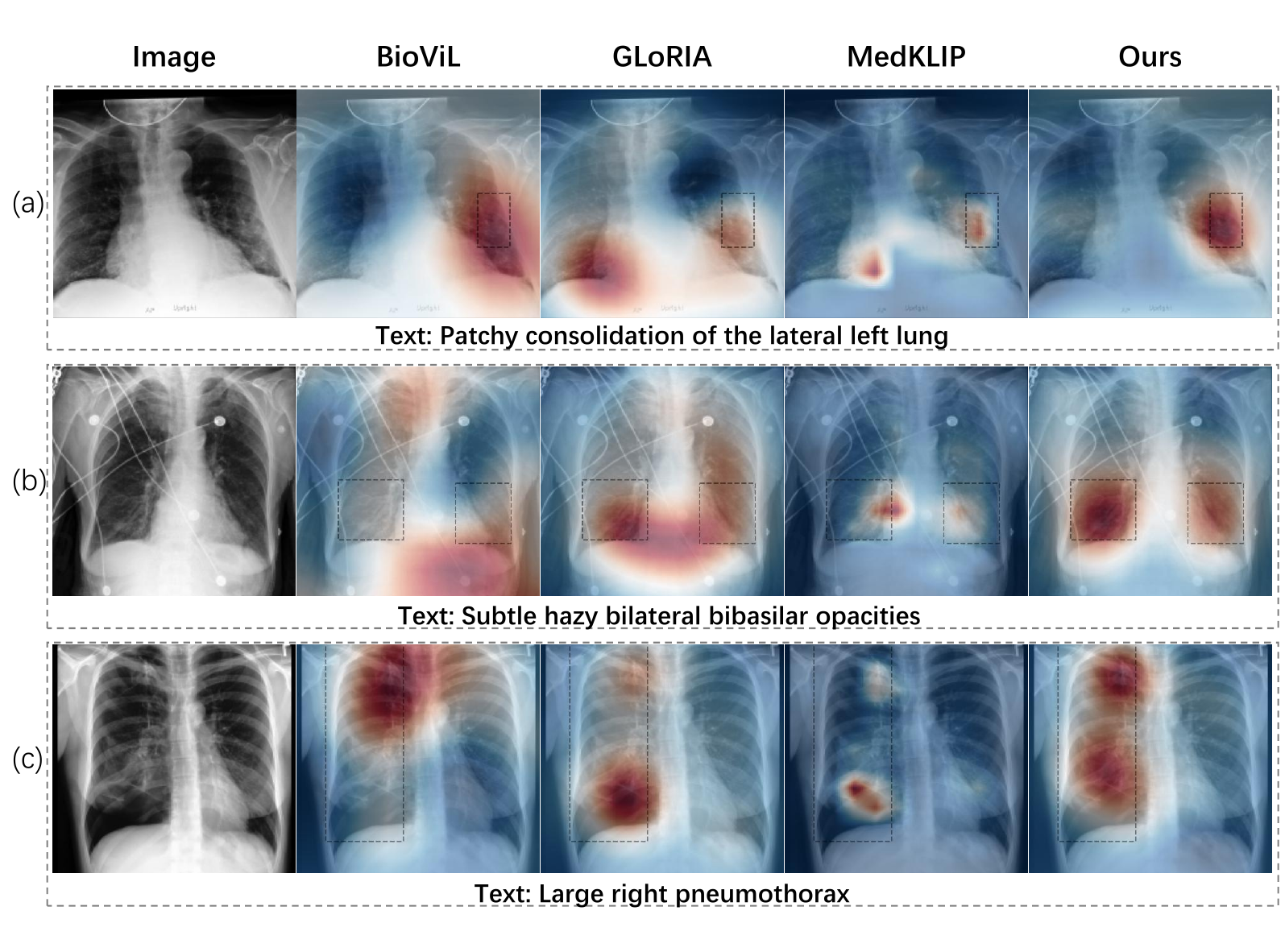}}
  \caption{Disease localization based on text descriptions. Subfigures a, b, and c display the results of disease localization for different pathological conditions.}
  \medskip
\end{figure}

For seen diseases, we present the results based on the RSNA and MS-CXR datasets in Table 1. Our proposed model outperforms existing methods on all metrics. For example, we improve the IOU score from 0.317 to 0.331 and the Dice score from 0.465 to 0.474. The experimental results on the MS-CXR dataset are shown in Tables 2 and 3, where our model performs exceptionally well for five pathological conditions, achieving the highest Dice and IOU scores. It also significantly surpasses existing methods in terms of the average Dice and IOU scores for eight pathological classes.

For unseen diseases, we also conducted localization experiments specifically for the novel coronavirus. As shown in Table 1, our model consistently demonstrates improvements across all metrics. We increase the IOU score from 0.137 to 0.222 and the Dice score from 0.228 to 0.336. These results indicate significant progress in disease localization achieved by our proposed model, performing well in identifying both seen and unseen diseases.

Our model excels in pathological localization as shown in Fig. 3, accurately identifying specific lesions from descriptions. For instance, Fig. 3a highlights our method's precise detection of a small "consolidation" lesion on the left lung, avoiding issues such as overly large or multiple false detections. In Fig. 3b, we accurately locate "lung opacities" where others struggle. Furthermore, Fig. 3c showcases our ability to detect a large pneumothorax, covering a broader area than competing methods. Our technique consistently outperforms others in localizing various lesion types.

\section{CONCLUSION}
\label{sec:conclu}
A novel multimodal self-supervised learning method has been developed aimed at precisely locating lesions in chest X-ray images through concise lesion descriptions. This approach does not rely on location annotations, but instead uses sentences from medical reports as semantic units to match local regions of the image. By integrating global and local contrastive losses, the method learns a shared latent semantic representation, thereby achieving fine-grained semantic alignment. To confirm the effectiveness of this method, it was validated on three independent external test datasets and achieved state-of-the-art results. The method demonstrates strong generalizability in handling various sizes and numbers of diseases. Moreover, it exhibits exceptional robustness when dealing with previously unseen diseases. This is of significant importance for advancements in the medical field, as the emergence of new diseases requires timely and accurate diagnosis and treatment plans. Overall, the method shows encouraging localization capabilities in disease detection, highlighting the immense potential of deep learning in medical diagnostic assistance with a large volume of unlabeled data.
\hfill
\pagebreak

\clearpage
\section{Compliance with ethical standards}
\label{sec:ethics}
This research study was conducted retrospectively using human subject data made available in open access. Ethical approval was not required as confirmed by the license attached with the open access data.

\section{Acknowledgments}
\label{sec:acknowledgments}
This research was partly supported by the National Natural Science Foundation of China (62222118, U22A2040), Shenzhen Science and Technology Program (RCYX20210706092-104034, JCYJ20220531100213029), Guangdong Provincial Key Laboratory of Artificial Intelligence in Medical Image Analysis and Application (2022B1212010011), the major key project of Peng Cheng Laboratory under grant PCL2023AS1-2, and Key Laboratory for Magnetic Resonance and Multimodality Imaging of Guangdong Province (2020B1212060051).
\setstretch{0.5}
\bibliographystyle{IEEEbib}
\bibliography{ISBI_latex}

\end{document}